\documentclass[letterpaper, 10 pt, conference]{ieeeconf}  % Comment this line out if you need a4paper

\IEEEoverridecommandlockouts                              % This command is only needed if 
\usepackage{graphics} % for pdf, bitmapped graphics files
\usepackage{url}
\usepackage{booktabs}
\usepackage{theorem}
\usepackage{amsmath} % assumes amsmath package installed
\usepackage{amsfonts}
\usepackage{amssymb}  % assumes amsmath package installed
\usepackage{multicol}
\usepackage{graphicx}
\usepackage{layouts}
\usepackage{wrapfig}
\usepackage{graphicx}
\usepackage{easyReview}
\usepackage{printlen}
\usepackage{bm}
\usepackage{siunitx}
\usepackage{colortbl}	% to color table background
\usepackage[utf8]{inputenc}
\usepackage{pgf}
\usepackage{tikz}
\usepackage{tikzscale}
\usepackage{pgfplots}
\DeclareUnicodeCharacter{2212}{\textendash}
\pgfplotsset{compat=1.14}
\usepgfplotslibrary{external}
\usepgfplotslibrary[external]
\tikzset{external/force remake}
\usepackage{algorithm}
\usepackage{algpseudocode}
\usepackage{makecell}
\usepackage{hyperref}
\usepackage[caption=false,font=footnotesize]{subfig}

\newtheorem{remark}{\textbf{Remark}}

\def\*#1{\mathbf{#1}}

\DeclareMathOperator*{\argmin}{arg\,min}

\title{\LARGE \bf
Trajectory Optimization under Contact Timing Uncertainties}
\author{Haizhou Zhao$^{1}$
 , Majid Khadiv$^{1}$ % <-this % stops a space
\thanks{$^{1}$Munich Institute of Robotics and Machine Intelligence, Technical
University of Munich, Germany {\tt\small haizhou\_zhao@outlook.com, majid.khadiv@tum.de}}
}

\begin{document}

\maketitle
\thispagestyle{empty}
\pagestyle{empty}

% \textcolor{red}{\textbf{TODO:} Images of robot hopping here (sim + real?).}

%%%%%%%%%%%%%%%%%%%%%%%%%%%%%%%%%%%%%%%%%%%%%%%%%%%%%%%%%%%%%%%%%%%%%%%%%%%%%%%%
\begin{abstract}
Most interesting problems in robotics (e.g., locomotion and manipulation) are realized through intermittent contact with the environment. Due to the perception and modeling errors, assuming an exact time for establishing contact with the environment is unrealistic. On the other hand, handling uncertainties in contact timing is notoriously difficult as it gives rise to either handling uncertain complementarity systems or solving combinatorial optimization problems at run-time. This work presents a novel optimal control formulation to find robust control policies under contact timing uncertainties. Our main novelty lies in casting the stochastic problem to a deterministic optimization over the uncertainty set that ensures robustness criterion satisfaction of candidate pre-contact states and optimizes for contact-relevant objectives. This way, we only need to solve a manageable standard nonlinear programming problem without complementarity constraints or combinatorial explosion. Our simulation results on multiple simplified locomotion and manipulation tasks demonstrate the robustness of our uncertainty-aware formulation compared to the nominal optimal control formulation.
\end{abstract}

%%%%%%%%%%%%%%%%%%%%%%%%%%%%%%%%%%%%%%%%%%%%%%%%%%%%%%%%%%%%%%%%%%%%%%%%%%%%%%%%
\section{Introduction}
Intermittent contact with the world renders locomotion and object manipulation problems hybrid. When using optimal control to generate plans for these systems, the resulting problem to solve would be a mixed-integer optimization problem \cite{deits2014footstep,toussaint2018differentiable}. Several works have tried to solve the problem by relaxing the hybrid nature, e.g., smoothing the contact transition by regularizing the Delasus matrix \cite{tassa2012synthesis}, handling physical consistency as a soft constraint \cite{mordatch2012discovery}, or relaxing contact with complementarity slackness in the solver \cite{posa2014direct}.
% While all of these approaches have shown impressive results in simulation, and some have become real-time capable for model predictive control (MPC) \cite{koenemann2015whole,neunert2018whole,dafarra2022dynamic}, their application in the real world has been very limited. There are many reasons for this, but the most important ones are: 1) the generated motions are highly sensitive to contact properties and do not transfer to the real world, 2) the highly nonlinear optimization problem with contact is prone to many local minima, most of which do not admit a desired behavior, and 3) the run-time computation is extremely large (and barely parallelizable) which results in slow control loops on real hardware. 
Most recent efforts to implement MPC for locomotion and manipulation have focused on solving a hierarchical problem instead of the holistic one and could achieve impressive behaviors on real hardware \cite{toussaint2022sequence, mastalli2023agile, grandia2023perceptive, meduri2023biconmp,zhu2023efficient}. These approaches consider a fixed contact plan and control the whole body motion for the given plan. 
%To handle the complexity of the whole-body MPC for multi-limp systems, they use a hierarchy of MPCs to make the controller more reactive (adapting the location and time of contact) \cite{li2021model,daneshmand2021variable}, separate kinematics from dynamics \cite{meduri2023biconmp}, apply an instantaneous inverse dynamics to track the generated motions \cite{grandia2023perceptive}, leverage a memory of motions \cite{dantec2021whole} or offline trajectories \cite{mastalli2023agile} to warm-start MPC. 
%
\begin{figure}[t]
    \centering
    \includegraphics[width=\linewidth]{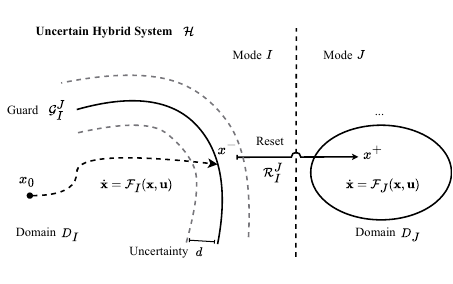}
    \caption{Illustration of Uncertain Hybrid Systems}
    \label{fig:uncertainhybrid}
\end{figure}
They also assume that contact events happen at exact times, i. e., the predefined switching times. However, in reality, this is a very restrictive assumption. For instance, the robot's perception of the environment is always with some errors. Furthermore, the tracking error of the end-effector establishing contact can also lead to a mismatch between the planned and realized time of contact.
To handle these situations, the whole-body MPC frameworks available in the literature either use heuristics \cite{grandia2023perceptive} or rely on the intrinsic robustness of MPC through fast replanning to handle uncertainties in contact events \cite{toussaint2022sequence,meduri2023biconmp,mastalli2023agile}. However, these approaches are very limited and a more systematic approach is required.

Recently, \cite{drnach2021robust,hammoud2021impedance,gazar2023nonlinear,gazar2023multi,SHIRAI2024101466} investigated the use of robust and stochastic optimal control for contact-rich robotics problems. While these approaches provide a very concrete understanding of the problem and interesting safety guarantees, they generally fall short in handling contact timing uncertainty. \cite{hammoud2021impedance} has shown that adjusting the end-effector impedance as a function of disturbances can mitigate the problem of impact when the contact event is uncertain. In this framework, the contact event is considered to be uncertain with a known distribution, and the impact is mitigated using a risk-sensitive optimal controller.
% In other words, formulating the control input as $u_t = k_{des,t} + K_t(x_{des,t}-x_t)$, risk-sensitive impedance profiles $K_t$  are found for a given desired trajectory of states $x_{des,t}$ and controls $k_{des,t}$ ($x_{t}$ is the measured state).
However, not adapting desired trajectories can limit the controller's capability in handling situations such as late foot touch-down during locomotion. 
\cite{Wang2020ImpactawareTQ} handles timing uncertainty by bounding the predicted post-impact states for safety through impact-aware task-space quadratic programming. However, the instantaneous nature of their controller may result in myopic behaviors and potential feasibility issues due to hard constraints. \cite{aaronconvergent,daisamplinglimitcircle} use sampling-based methods to consider both post-impact safety and feasibility. However, these approaches do not scale to full manipulation and locomotion problems.

The primary contribution of this work is to provide a deterministic re-formulation of the stochastic hybrid optimal control problem with uncertainty in the switching event that does not add run-time computational complexity compared to the deterministic optimal control problem. In doing so, our formulation adds a robust phase to the problem that accounts for a trajectory of possible switching states over the uncertainty set. The proposed approach can be adapted for general contact dynamics and is applicable to both locomotion to manipulation problems. Through several simplified examples, we demonstrate the robustness of our approach compared to the standard nominal optimal control problem. 

The rest of the paper is structured as follows: in Section \ref{sec:preliminaries}, we provide the necessary ingredients to formulate the problem. In section \ref{sec:approach}, we detail our proposed formulation. In section \ref{sec:case_study}, we present the results of applying our formulation to several simplified locomotion and manipulation problems. Finally, Section \ref{sec:conclusions} presents the concluding remarks and future work.
% Recently, \cite{laouar2023feasibility} proposed an approach named consensus-horizon MPC that takes the uncertainty in switching surfaces into account to enable a robust solution. Here, we extend this algorithm and propose a consensus-horizon MPC that can handle uncertainties in the contact event.

\section{Preliminaries}\label{sec:preliminaries}
In this section, we first define the terminology required for describing our problem. Then, we present a deterministic optimal control formulation for our hybrid dynamical system.
\subsection{Deterministic Hybrid Systems}\label{sec:prelim_hybridsys}
Locomotion and manipulation are realized through intermittent contact with the environment. One way to formalize this problem is through the framework of hybrid dynamical systems \cite{westervelt2003hybrid}. In this work, we consider the following definition of hybrid systems \cite{aaron2016ijrr_hybrid,hybridilqr2023aaronkong}
\begin{equation}\label{eq:hybridsystem}
    \mathcal{H}:\bigg\{\begin{array}{ll}
    \dot{\*x}=\mathcal{F}_I(\*x,\*u),&\*x\in\mathcal{D}_I\backslash\mathcal{G}_I^J,\*u\in\mathcal{U}_I,\\
    \*x^+=\mathcal{R}_I^J(\*x^-),&\*x^-\in\mathcal{G}_I^J,\*x^+\in\mathcal{D}_J,\\
    \end{array}
\end{equation}
with $\mathcal{J}=\{I, J, ...\}$ being the finite set of discrete modes such that for a mode $I \in \mathcal{J}$,
\begin{itemize}
    \item $\mathcal{F}_I$ is the continuous dynamics,
    \item $\mathcal{D}_I$ is the domain of states,
    \item $\mathcal{U}_I$ is the set of admissible control inputs,
    \item $\mathcal{G}_I^J:=\{\*x\in\mathcal{D}_I|g_I^J(\*x)=0,\dot{g}_I^J(\*x)\le0\}$ is the guard (Fig. \ref{fig:uncertainhybrid}),
    \item $\mathcal{R}_I^J:\mathcal{G}_I^J\to\mathcal{D}_J$ is the reset map that projects states in $D_I$ to $D_J$ when the guard condition $g_I^J(\*x^-)\le0$ is met.
\end{itemize}
A simple example of a hybrid robotic system is a jumping 1D hopper. Upon landing, the robot's states enter the guard from the aerial phase to stance, undergoing a reset by an impulsive impact force.
\subsection{From Hybrid to Switching systems}
Given the sequence of contacts for a hybrid system, the problem can be simplified to a switching system. In this formulation, the system's dynamics are smooth between consecutive switches, while the time of the switch can still be optimized. Recently, many fast solvers \cite{OCS2,crocoddyl,crocoddylnullspace} have been developed for real-time resolution of \eqref{eq:timebasedoc}. In the following, we present the multiple-shooting transcription of the switching system.

Let $\mathcal{S}$ be the set of shooting node indices where a switch is expected. For a given initial state $\*x_0$, the time-based direct-multiple-shooting optimal control problem can be formulated as
\begin{subequations}\label{eq:timebasedoc}
    \begin{align}
    \min_{\*x,\*u}\quad&L_N(\*x_N)+\sum_{i=0}^{N-1}L_i(\*x_i,\*u_i)\\
    \text{s.t.}\quad&\forall i \in[0, N-1],\*h_i(\*x_i,\*u_i)\le0\label{eq:ineqnormal},\\
    &\*h_N(\*x_N)\le0,\label{eq:ineqnormal_terminal}\\
    &\forall i\notin\mathcal{S}:\nonumber\\
    &\quad\*f_i(\*x_{i},\*u_i,\*x_{i+1}, \Delta t_i)=\*0,\label{eq:impdyn}\\
    &\quad g_i(\*x_{i+1})>0,\label{eq:guardpreineq}\\
    &\forall i\in\mathcal{S}:\nonumber\\
    &\quad\*f_i(\*x_{i},\*u_i,\*x_{i+1}^-,\Delta t_i)=\*0,\label{eq:impdynpre}\\
    &\quad\*x_{i+1}=\mathcal{R}_i(\*x_{i+1}^-),\label{eq:resetoc}\\
    &\quad g_i(\*x_{i+1}^-)=0,\label{eq:guardpreeq}
    \end{align}
\end{subequations}
where $\Delta t_i$ is the phase-wise timestep, $N$ is the number of shooting nodes, $L_N$ is the terminal cost, $L_i$ is the running cost, \eqref{eq:ineqnormal} is the state-input inequality constraints, \eqref{eq:ineqnormal_terminal} is the terminal state inequality constraints, \eqref{eq:impdyn} is the non-switching implicit dynamics, $\eqref{eq:impdynpre}$ is the pre-switching continuous dynamics derived from $\mathcal{F}_{(\cdot)}$ in \eqref{eq:hybridsystem}, $x^-_{i+1}$ denotes the pre-reset state, \eqref{eq:guardpreineq},\eqref{eq:guardpreeq} ensure switching consistency, and \eqref{eq:resetoc} is the state reset equation at the switch. 
\section{Uncertainty-Aware Optimal Control}\label{sec:approach}
The formulation in \eqref{eq:timebasedoc} assumes that contact happens at a certain time and state (where the distance between the end-effector and the environment goes to zero). However, due to uncertainties in the environment perception and end-effector tracking errors, it is highly unlikely that the end-effector touches the ground at the exact pre-defined time. to formalize this situation, 
we introduce the following uncertain guard as illustrated in Fig. \ref{fig:uncertainhybrid}:
\begin{equation}\label{eq:uncertainguard}
    \hat{\mathcal{G}}_I^J(\delta)=\{\*x\in\mathcal{D}_I|{g}_I^J(\*x)= \delta,\dot{g}_I^J(\*x)\le0\},\delta\in[-d,d],
\end{equation}
where $\delta$ is the guard uncertainty bounded by $d$. With $\hat{\mathcal{G}}$, a state cannot be deterministically predicted to incur switching, leading to uncertain contact timing and thus the switching time between modes. This is naturally incompatible with the deterministic structure of \eqref{eq:timebasedoc}. 
\subsection{Issues of the Nominal Approach}\label{sec:nominal_approach}
Trajectories generated from nominal time-based optimal control with a nominal guard ($\delta=0$) only ensure that the nominal switching state is feasible. If the switching does not happen as planned (i.e., early or late contact), the system may evolve unexpectedly, leading to the following issues:
\begin{itemize}
    \item For late contact, the designed controller is not valid after the nominal contact timing due to the mode mismatch between the reference and the actual trajectories. Problem-specific solutions include reference spreading \cite{saccon2024tro}, simplistic zero-order-hold of the last input, and event-triggered controllers that formulate a locomotion task as a finite state machine \cite{raibertreport}.
    \item For early contact, the system usually encounters unfavorable impact forces. In such cases, the system may fail due to insufficient mechanical strength or bounce of the end-effector, e.g., feet of legged robots.
    \item Since the nominal problem assumes an exact contact event, it can lead to highly aggressive motions before or after contact. A consequence is that the actual switching states may fall outside the feasibility set of the post-impact problem.
\end{itemize}
In the rest of this section, we introduce our main contribution: an uncertainty-aware optimal control formulation that resolves the above issues.
% Using stochastic formulation may avoid explicit handling with the uncertain guard. However, it does not provide a safety or feasibility guarantee within an uncertainty set if only the expectation  
% Using a nominal guard or a stochastic formulation in optimal control will generate potentially infeasible trajectories since switching might occur at any time close to the nominal switching time.
\subsection{A Deterministic Transcription}
\begin{figure}[t]
    \centering
    \includegraphics[width=\linewidth]{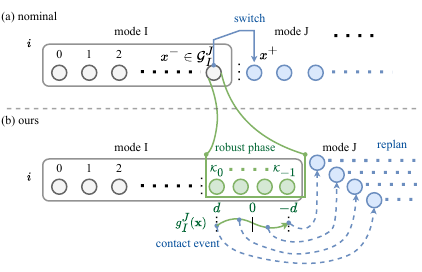}
    \caption{Illustration of the difference between (a) the nominal optimal control and (b) the proposed approach. The proposed method does not switch the mode of the states within the robust phase, but generates a trajectory of feasible pre-switch states over the uncertainty set.}
    \label{fig:formulation_illustration}
\end{figure}
Intuitively, a robust time-based trajectory that solves the problem in Sec. \ref{sec:nominal_approach} should be deterministic until a switch is triggered. Such contact events can be detected in various ways such as contact sensors. When the states are within the uncertain region, all candidate pre-switching states should result in a feasible and safe motion.
Based on this intuition, we consider an uncertain sub-phase (namely the \textbf{\textit{robust phase}}), appended to the pre-switching phase. An illustration can be found in Fig. \ref{fig:formulation_illustration}. Let $\mathcal{K}$ be the index set of the robust phase shooting nodes. For the robust phase, the following constraints must be satisfied:
\begin{subequations}\label{eq:uncertainconstr}
\begin{align}
    i=\mathcal{K}_0,~&g_i(\*x_i)=d,\\
    i=\mathcal{K}_{-1},~&g_i(\*x_i)=-d,\\
\forall i\in \mathcal{K},~&\dot{g}_i(\*x_i)\le0,\label{eq:uncertainconstr_monotonicity}
 % \; \& \; i \neq 0,-1
\end{align}
\end{subequations}
where $\mathcal{K}_0,\mathcal{K}_{-1}$ denote the first and last indices in $\mathcal{K}$, respectively. \eqref{eq:uncertainconstr} ensures that the trajectory traverses the uncertain region, constituting a collection of possible switching states, where $\eqref{eq:uncertainconstr_monotonicity}$ ensures the satisfaction of the time-derivative part of \eqref{eq:uncertainguard}. Note that reset map \eqref{eq:resetoc} is not applied to any pre-switch states within the robust phase. For a single phase, an uncertainty-aware optimal control problem can then be formulated as a parameterized optimization problem as introduced in \cite{Oshin2022ParameterizedDD}:
\begin{subequations}\label{eq:uncertainoc}
    \begin{align}
    \min_{\*x,\*u,\Delta t,d}\quad&\sum_{i=0}^{N-1}L_i(\*x_i,\*u_i)+\sum_{i\in\mathcal{K}}L_K(\*x_i, \*p_i)\\
    \text{s.t.}\quad
    &\forall i\in[0,N-1]\cup\mathcal{K}:\eqref{eq:impdyn}, \nonumber\\
    &\quad\Delta t_i\in[\Delta t_{\min},\Delta t_{\max}]\label{eq:timebounds}\\
    &\forall i\in[0,N-1]: \eqref{eq:ineqnormal}\nonumber\\
    &\forall i\in\mathcal{K}:\eqref{eq:uncertainconstr},\nonumber\\
    &\quad\*h_K(\*x_i, \*p_i)\le\*0,\label{eq:uncertainfeasibleset}\\
    &\forall i\notin\mathcal{K}: \eqref{eq:guardpreineq}\nonumber\\
    &d\in[d_{\min},d_{\max}],\label{eq:uncerbounds}
    \end{align}
\end{subequations}
where $\mathcal{K}_0=N$, $L_K$ is the contact-related objective, $\*h_K$ is the contact-related constraint, and $\*p_i$ is the collection of auxiliary variables including the timesteps and uncertainty. $\Delta t$ and $d$ are decision variables in this new formulation, bounded by \eqref{eq:timebounds} and \eqref{eq:uncerbounds} where $(\cdot)_{{\min}/{\max}}$ stand for lower and upper bounds repectively. They are crucial for ensuring the feasibility and convergence of the optimization problem.

By incorporating the robust phase in the form of \eqref{eq:uncertainconstr}, the proposed formulation actively takes into account the effect of uncertain switching surface using $\*h_K,L_K$, in the following ways: we can model safety-related or feasibility-related criteria as inequality constraints $\*h_K$, e.g., adding safety constraints on predicted post-impact states \cite{Wang2020ImpactawareTQ} or a viability kernel for stability \cite{troviabilitybipedal}; we can also adapt $L_K$ to reach various goals such as finding the maximum uncertainty of ground height such that a bipedal robot can remain stable and the minimal impulse a jumping robot can achieve. Compared to \eqref{eq:timebasedoc}, our formulation makes the following modifications; in the case of early contacts, the pre-switch robustness ensures the states will not cause large impacts or infeasibility; In the case of late contacts, the robust phase solution makes sure the system is feasible for post-impact replanning. In the next section, we will show the flexibility of our formulation in different case studies.

\begin{remark}
    (Uncertainty optimization) The proposed formulation \eqref{eq:uncertainoc} finds a single trajectory to cover all possibilities within the uncertainty region parameterized by $d$. If $d$ is too large, the solution might be too conservative to satisfy constraints or minimize costs for the worst case. If $d$ is too small, the solution will be too aggressive. Therefore, in our formulation, by adjusting the cost function for different problems,  $d$ can be optimized for feasibility and the trade-off between conservativeness and aggressiveness.
\end{remark}
\begin{remark} (Optimality)
As a single-phase formulation, the uncertain-aware formulation \eqref{eq:uncertainoc} cannot handle the long-term optimality as opposed to the nominal formulation in \eqref{eq:timebasedoc}. Nevertheless, it can further be extended to a parallelizable tree-structured optimal control problem \cite{hpipm} that branches at each shooting node in $\mathcal{K}$. In this paper, we only focus on the transcription of the uncertainty into a robust phase and leave the long-term optimality problem as future work.
\end{remark}
\section{Case Studies}\label{sec:case_study}
In this section, we show case studies of various locomotion and manipulation tasks based on the proposed optimal control formulation. We also compare the results of our proposed robust formulation to the nominal case. All examples are implemented using the Opti stack of CasADi \cite{casadi} and IPOPT \cite{Wchter2006ipopt}. Simulation and result visualization can be found in https://www.youtube.com/watch?v=4z9JyT7jI-w.

\subsection{Impact Minimization of a Hopping Robot}\label{sec:hopping}

\begin{figure}[t]
    \centering
    \includegraphics[width=\linewidth]{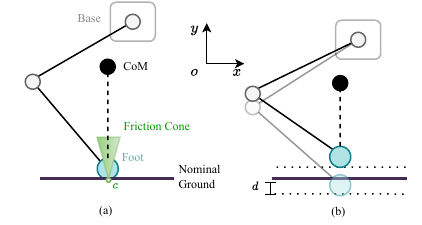}
    \caption{Illustration of the planar two-link point-footed robot. (a) The robot has two joints (hip and knee) and a 2-DoF base. The black dot denotes the CoM. (b) When landing, the ground position is uncertain.}
    \label{fig:hopper}
\end{figure}

\begin{figure*}[t]
    \centering
    \includegraphics[width=\linewidth]{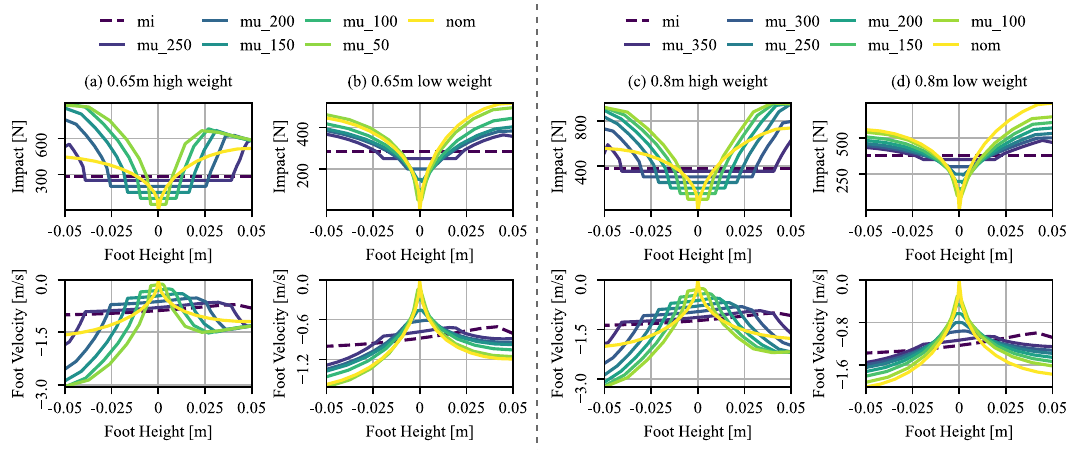}
    \caption{Simulation data during the robust phase. The weights for maximizing the uncertainty in (a),(c) are respectively 1000x that in (b),(d). 'mi' denotes the impact minimization over the given uncertainty [-0.05, 0.05]m. 'mu\_(x)' denotes uncertainty maximization for known impact limits x (unit: N), where the flat region denotes the uncertainty. 'nom' denotes the nominal optimal control data. Flat parts of 'mu\_(x)' denote the optimized uncertainty region where the impact limits are satisfied.}
    \label{fig:0.8data}
\end{figure*}
One of the classical examples in robot locomotion control is impact minimization for jumping robots \cite{bogdanovic2020learning}. In this task, a planar two-link point-feet hopper jumps continuously while the height of the support surface can suddenly change within a bound, as shown in Fig. \ref{fig:hopper}. The robot has a 2-DoF X-Y base joint.
The task is to perform in-place hopping to reach a desired height. 
\subsubsection{Dynamic Model} 
Let $\*{q}=[x_b, y_b, \theta_h, \theta_k]^\top$, and $\dot{\*q}=[ \dot{x}_b, \dot{y}_b,\dot{\theta}_h,\dot{\theta}_k]^\top$. $y_b$ is the base height, and $\theta_h,\theta_k$ are the hip and knee angles, respectively. The dynamics of the system can be written as
\begin{subequations}
    \begin{align}
    \*M(\*q)\ddot{\*q}+\*H(\*q,\dot{\*q})&=\*S\boldsymbol{\tau}+\*J_c^\top\*F_c, \, \label{eq:hopperdynamics}
    % \*J_c\ddot{\*q}+\dot{\*J}_c\dot{\*q}&=\*0,\*J_c\dot{\*q}=\*0,\label{eq:hopperpointcontact}
    \end{align}
\end{subequations}
where 
% \eqref{eq:hopperdynamics} is the equation of motion, \eqref{eq:hopperpointcontact} is the point contact constraint when the foot is in contact,
$\*M\in\mathbb{R}^{4\times4}$ is the joint-space inertia matrix, $\*H\in\mathbb{R}^4$ is the nonlinear effects, $\*S=\begin{bmatrix}\*0_{2\times1}&\*I_2\end{bmatrix}^\top$ is the selection matrix, $\boldsymbol{\tau}\in\mathbb{R}^2$ is the joint torques, $\*J_c\in\mathbb{R}^{2\times4}$ is the foot contact jacobian, $\*F_c=[F_y,F_x]^\top\in\mathbb{R}^2$ is the contact force subject to the following constraints:
\begin{subequations}
    \begin{align}
        0<F_y &~\bot~y_f - y_g>0, \label{eq:hoppercomplementary}\\
        \mu F_y &\ge|F_x|,\label{eq:hopperfrictioncone}
    \end{align}
\end{subequations}
where \eqref{eq:hoppercomplementary} is the contact complementary constraints, $y_f,y_g$ are the foot and the ground height, respectively. Equation \eqref{eq:hopperfrictioncone} encodes the planar friction cone constraint. In the trajectory optimization, we assume purely inelastic impact, i.e., zero post-impact foot velocity. Based on the maximum dissipation principle, the impact impulse $\boldsymbol{\lambda}=[\lambda_x,\lambda_y]^\top$ can be modeled as
\begin{subequations}\label{eq:hopperimpulse}
    \begin{align}
        \boldsymbol{\lambda}=&\argmin~||\*J_c^\top\dot{\*q}^+||^2\\
        \text{s.t.}\quad&\*M(\dot{\*q}^+-\dot{\*q}^-)=\*J_c^\top\boldsymbol{\lambda}+[\*S\boldsymbol{\tau}-\*H(\*q,\dot{\*q}^-)] \Delta t,\\
        &\mu\lambda_y\ge|\lambda_x|,\\
        &\*J_c\dot{\*q}^+=0,
    \end{align}
\end{subequations}
where $\Delta t$ denotes the impact duration, which is set to be 2 ms in our tests. Note that impulse is used instead of force to improve the numerical conditioning. 
\subsubsection{Nominal Optimal Control}
In the form of \eqref{eq:timebasedoc}, a hopping loop is divided into three phases: take-off (stance), ascendance, and falling. A terminal constraint of the base height is added to the ascendance phase to ensure the base reaches the desired position. The guard is chosen as
\begin{equation}\label{eq:hopperguard}
    g_i:=y_f-y_g.
\end{equation}
Let $\*r_f=[x_f,y_f]^\top$.
To maintain the discretized contact constraint during the stance phase, we add the velocity-level stabilization at each shooting node:
\begin{equation}
    k_f\dot{\*r}_f+\*r_f=\*r_f^0,
\end{equation}
where $k_f=1e3$ in our setting, $\*r_f^0$ is the initial foot position. The center-of-mass (CoM) of the robot is set to be right above the foot during the whole procedure for in-place hopping. Upon switching, \eqref{eq:hopperimpulse} is added and the horizontal post-impact velocity of the foot is constrained to be zero as a terminal constraint of the falling phase to avoid slip. Torques, joint positions, and velocities are also constrained with realistic values.

\subsubsection{Robust Formulation}
The robust formulation is the same as the nominal one except that an extra robust phase is added. The guard \eqref{eq:hopperguard} is used in the form of \eqref{eq:uncertainconstr}. Two realistic scenarios are tested to show the flexibility of our method:
\begin{itemize}
    \item Minimizing impact force for the worst-case uncertainty within a given range. In this case, we minimize the upper bound of vertical impact $\bar{\lambda}_y$ in the robust phase, i.e.,
    \begin{subequations}
        \begin{align}
        L_K&:=w_\lambda\bar{\lambda}_y,\\
        h_K&:=\lambda_y<\bar{\lambda}_y.
        \end{align}
    \end{subequations}
    In this case, $d$ is a parameter and $\bar{\lambda}_y$ is a decision variable.
    \item Maximizing uncertainty based on the worst-feasible impact force. This is the safety-critical case when the maximum tolerable impact by the structure of the robot $\bar{\lambda}_y$ is obtained from mechanical design. In this case, for the robust phase, we have: 
    \begin{subequations}
        \begin{align}
        L_K&:=-w_dd,\\
        h_K&:=\lambda_y<\bar{\lambda}_y.
        \end{align}
    \end{subequations}
    In this case, $\bar{\lambda}_y$ is a parameter and $d$ is a decision variable.
\end{itemize}
\subsubsection{Result and Discussion} Two desired heights (0.65m, 0.8m) are tested for the nominal approach and the two scenarios of the robust approach. The friction coefficient is set at 0.7. In terms of impact minimization, it can be observed in Fig. \ref{fig:0.8data} that the robust method can have approximately up to 30\%-50\% improvement over the nominal method for about 70\% of the uncertain region.

For uncertainty maximization, a wide range for the weight $w_d$ is considered to generate diverse solutions. For low $w_d$, as the impact limit $\bar\lambda \to 0$, the robust solution converges to the nominal case. For high $w_d$, the feasible uncertainty can be larger, at the cost of higher impact force outside the uncertain region. The low $w_d$ cases can also be interpreted as reducing the uncertainty to obtain better average improvements over the nominal method i.e., the percentage of the original uncertain region with lower impact forces than the nominal solution. We also compared the robust method with the nominal in a real-time control scenario by using the off-the-shelf simulator RaiSim \cite{hwangbo2018per}, for which one can refer to the supplementary video.
% \subsubsection{Online tracking of generated trajectories} We test the trackability of the trajectories generated from the proposed method in RaiSim when changing the ground height for each loop, to reproduce the hopping behavior in \cite{bogdanovic2020learning}. \textbf{TODO}
\subsection{Object Catching}
\begin{figure}[t!]
    \centering
    \includegraphics[width=\linewidth]{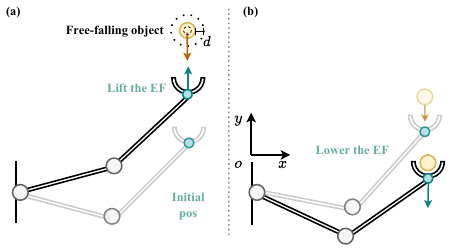}
    \caption{Illustration of how the manipulator catches a free-falling object. The manipulator (a) lifts its end-effector (EF) to a high position and then (b) lowers its EF to reduce the velocity w.r.t. the object.}
    \label{fig:objectcatching}
\end{figure}
\begin{figure}[t!]
    \centering
    \begin{minipage}[t]{\linewidth}
    \includegraphics[width=\linewidth]{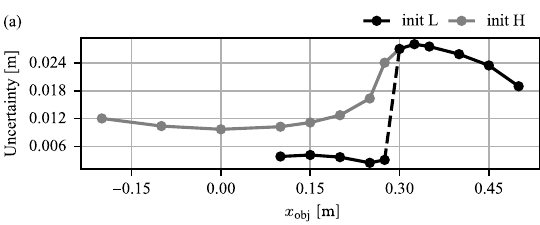}\end{minipage}
    \begin{minipage}[t]{\linewidth}
    \includegraphics[width=\linewidth]{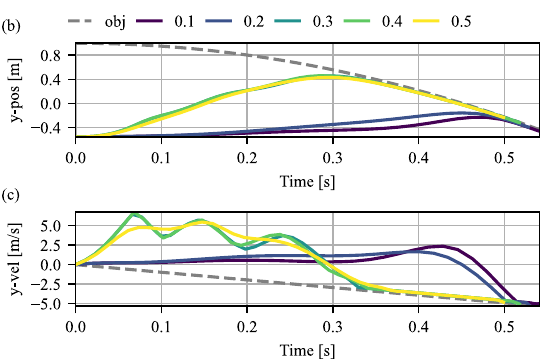}
    \end{minipage}
    \caption{Optimization and simulation data of the manipulator object-catching task. (a) The (solved) optimized uncertainty w.r.t. the initial $x_\text{obj}$ with differential initializations. (b),(c) are the y-position and -velocity trajectories of the object and the EF with the 'init L' initialization, where the number denotes the $x_\text{obj}$. The manipulator follows the strategy of reducing velocity difference at possible impacts.}
    \label{fig:objectcatchingdata}
\end{figure}
The second study is a torque-controlled 2DoF robotic arm catching an object, of which the shape is uncertain. It is a typical safety-critical case as an object can be fragile and may break if the impact upon contact is high. The setup is shown in Fig. \ref{fig:objectcatching}. For simplicity, instead of using impulse as safety criteria, it is assumed that the object is damaged if the impact velocity difference between the end-effector (EF) and the object exceeds a maximal value.

\subsubsection{Dynamic Model}
The arm shoulder joint (the first joint attached to the fixed base) is located at the origin. The inverse dynamics of the arm can be derived by the Recursive Newton-Euler Algorithm (RNEA) in its joint space. The object is underactuated and falls freely.

\subsubsection{Experiment Setting}
In this experiment, the robust method is not compared with the nominal approach since the result will be similar to Sec. \ref{sec:hopping}, i.e., too high impacts at unexpected contacts that can break the object. Instead, we show how the robust method can deal with the time-varying contact surface of the falling object.

Let $y_\text{object},y_\text{EF}$ be respectively the y-position of the nominal bottom of the object and the EF of the arm. The uncertain guard in \eqref{eq:uncertainconstr} is chosen as
\begin{equation}
    g_i:=y_\text{obj} - y_\text{EF},
\end{equation}
with the following constraints on the x-positions of the object and the EF to ensure consistent geometry during catching
\begin{equation}
    \forall i\in\mathcal{K}, x_\text{EF} = x_\text{obj}, \dot{x}_\text{EF}=0.
\end{equation}
The object falls from $y_\text{obj}=1$m and different $x_\text{obj}$. For all test cases, the arm starts from the same pose and zero velocity. The effect of the problem nonconvexity on the robust method was tested by using the following different optimizer initializations:
\begin{itemize}
    \item 'init L': initialization using the initial state where $y_\text{EF}<0$, i.e., the EF is lower than the shoulder joint.
    \item 'init H': initialization using the state where $y_\text{EF}>0$, i.e., the EF is higher than the shoulder joint.
\end{itemize}
\subsubsection{Result and Discussion}
As shown in Fig. \ref{fig:objectcatchingdata}(a), solutions to the two initializations diverge from each other for $x_\text{obj}\lessapprox0.28$ as represented by the discontinuity (black dashed line). Optimization with 'init L' is infeasible with low $x_\text{obj}$. In Fig. \ref{fig:objectcatching}(b,c), for 'init L', the y-position and velocity trajectories with $x_\text{obj}\in\{0.1, 0.2\}$ are drastically different from the ones with $x_\text{obj}\in\{0.3, 0.4, 0.5\}$. These indicate that the uncertainty optimization is affected by the non-convexity of the problem.  Interestingly, as can be seen in \ref{fig:objectcatching}(b,c), the manipulator reduces the velocity difference between its EF and the object to reduce the impact force. 
\subsection{Cart-Pole With a Rigid Wall}
\begin{figure}[t]
    \centering
    \includegraphics[width=\linewidth]{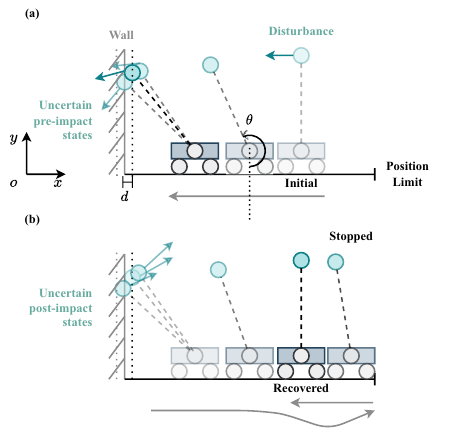}
    \caption{Illustration of the cart-pole system recovering balance. (a) The pole angular velocity is disturbed. Since the cart input is limited, it moves to the wall to seek impact that will reverse the direction of pole velocity. (b) After Impact, the cart-pole can recover its balance and position.}
    \label{fig:cartpole}
\end{figure}
In this case, we test a task similar to \cite{alp2022admm} where a cart-pole system can use contact with the wall to stabilize itself under disturbance. As shown in Fig. \ref{fig:cartpole}, the pole will bounce when colliding with a rigid wall, and the cart position is limited.
% It is expected to utilize the wall to recover its balance and return to the desired position. 
\subsubsection{Dynamic Model} Let $\*q=[x, \theta]$ where $x$ is the cart position and $\theta$ is the counterclockwise pole angle. Similar to the hopping robot, the equations of motion of the cart-pole can be written as
\begin{subequations}
    \begin{align}
        &\*M(\*q)\ddot{\*q}+\*H(\*q,\dot{\*q})=[1, 0]^\top\tau+\*J_c^\top\*F_c\\
        &\*M=\begin{bmatrix}
            m_c+m_p& m_plc_\theta\\
            m_plc_\theta&m_pl^2
        \end{bmatrix},\*H=-m_pls_\theta\begin{bmatrix}
            \dot{\theta}^2\\
            g
        \end{bmatrix},
    \end{align}
\end{subequations}
where $m_c,m_p$ are respectively the mass of the cart and the pole, $l$ is the pole length (its CoM is assumed to be at the end), $c_\theta, s_\theta$ are cosine and sine of the pole angle, $\tau$ is the cart linear driving force, $\*J_c\in\mathbb{R}^{2\times2}$ is the contact jacobian of the pole and $\*F_c\in\mathbb{R}^2$ is the wall reaction force. When the pole is upright, $\theta=\pi$. Its impact model is similar to \eqref{eq:hopperimpulse} except that for the normal velocity w.r.t. the wall $v_N$, we assume a restitution coefficient $C$, such that
\begin{equation}\label{eq:restitution}
    v_N^+=-Cv_N^-.
\end{equation}
\subsubsection{Nominal Optimal Control} The nominal optimal control comprises the pre-impact and post-impact phases. The pole is constrained to collide with the wall at the terminal node of the pre-impact phase. In the case of early contact, the nominal optimal control becomes a single-phase problem with the post-impact states as its initial state. For late contact cases, the wall position is updated to the actual value if the nominal contact is not triggered. 
\subsubsection{Robust Optimal Control}
\begin{figure}[t]
    \centering
    \subfloat[Convex hull]{
    \includegraphics[width=.47\linewidth]{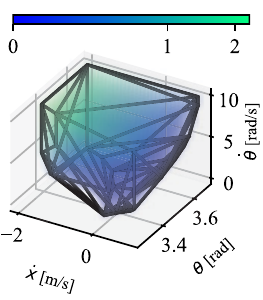}
    \label{fig:cartpoleconvexhull}}\hfil%
    \subfloat[Fitting error (m)]{
    \includegraphics[width=.47\linewidth]{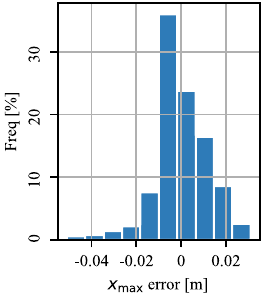}
    \label{fig:cartpoledistance}}
    \caption{Approximation of the feasible set of the cart-pole system. The colors represent (a) the stopping distance (unit: m) and (b) the fitting error of the quadratic approximation tested on the sampled data.}
\end{figure}
\begin{figure}[t]
    \centering
    \includegraphics[width=\linewidth]{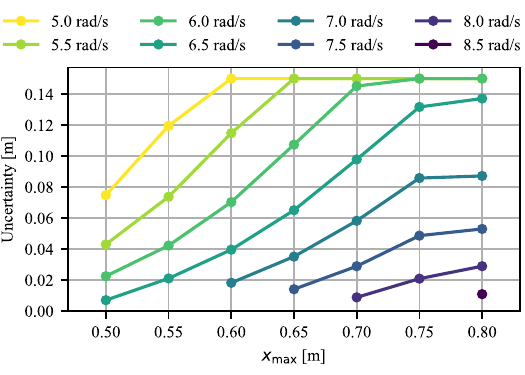}
    \caption{Optimized uncertainty for difference disturbed angular velocities $\dot{\theta}(0)$  and position limits $x_{\max}$. Only feasible solutions are plotted. }
    \label{fig:cartpoleuncert}
\end{figure}

% Since our proposed method contains only a single phase and the cart-pole is a non-minimum phase system, it is important to compute a feasible set of pre-switching states that ensure the feasibility of the optimization problem and the stability of the system.
Since the cart-pole is an unstable and constrained system, it is important that the robot's state after the impact remains in a set from which there exists a solution to stabilize the system under the constraints (a.k.a viability). In general, finding this set is very difficult and out of the scope of this paper. Here, we present a simple brute-force approach to approximate this set.
We used grid search to sample a small batch of pre-impact states $\*x=[\dot{x},\theta,\dot{\theta}]^\top$ and approximated the feasible ones as a convex hull as shown in Fig. \ref{fig:cartpoleconvexhull}. The stopping distance, i.e., the maximum position of the cart during the balancing, is approximated by a quadratic function $\phi$ of the pre-impact states as shown in Fig. \ref{fig:cartpoledistance}. These two approximations are sufficient for robust optimization with different $x_{\max}$.

Let $\*A\*x+\*b\le\*0$ be the convex hull. The constraints of the robust phase can be designed as
\begin{subequations}
    \begin{align}\
        \*h_K^\text{cvxh}&:=\*A\*x+\*b+\*s,\label{eq:carpole_feasibility_convexhull}\\
        h_K^\text{dist}&:=\phi(\*x)-x_{\max},\label{eq:cartpole_distance}
    \end{align}
\end{subequations}
where $\*s$ is the conservativeness parameter, \eqref{eq:carpole_feasibility_convexhull} is the convex hull constraint and \eqref{eq:cartpole_distance} is the maximum stopping distance constraint.
\begin{remark}
    (Conservativeness) As an outer linear approximation to the feasible set through sampling, the convex hull does not theoretically ensure feasibility, and states close to the boundary may be infeasible. To approximate an inner fit to the real feasible set, the conservativeness parameter $\*s$ is added to shrink the boundary. 
\end{remark}
\subsubsection{Results and Discussion}

\begin{figure}[t]
    \centering
    \includegraphics[width=\linewidth]{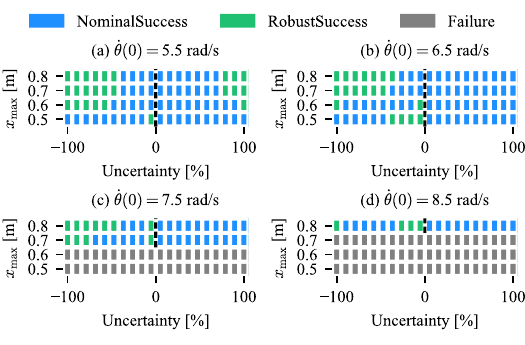}
    \caption{Success-failure plot for the comparison experiment. The x-axis of each subplot denotes the tested $\delta$ represented as a percentile of the optimized uncertainty (i.e., $\delta/d$) by the robust method w.r.t. the $x_{\max}$ and the perturbed $\dot{\theta}(0)$. 'NominalSuccess' denotes the success achieved by purely the nominal approach. 'RobustSuccess' denotes the success achieved by the robust method in addition to the 'NominalSuccess'. Note that the robust method will also succeed in 'NominalSuccess' settings. The dashed lines denote that the nominal method can find a nominal solution for the given setting (zero uncertainty).}
    \label{fig:cartpoleblock}
\end{figure}
The robust methods are tested on various $\dot{\theta}(0)$ and $x_{\max}$ settings. The restitution coefficient in \eqref{eq:restitution} is 0.8 and the friction coefficient is 0.7. The optimized uncertainties are shown in Fig. \ref{fig:cartpoleuncert} where the monotonicity w.r.t. $\dot{\theta}(0)$ and $x_{\max}$ can be summarized as respectively negative and positive.

Using the optimized uncertainty, we conducted further comparisons between the proposed method and the nominal approach to check its robustness, as shown in Fig. \ref{fig:cartpoleblock}. As can be seen, the robust method has a higher success rate for both early and late contacts. 
Taking the case of $\dot{\theta}(0)=5.5$ rad/s, for example, the robust method can succeed in all cases within the optimized uncertainty; in contrast, the nominal approach can only rely on the inherent margin of feasibility of the problem, and fail for too-early or too-late contacts, such as the failure for $x_{\max}=0.7$m with $\delta/d$ lower than $-40\%$ or higher than $70\%$.
This indicates that the robustness of the nominal approach is limited by its potentially aggressive solution. Nevertheless, the robust approach cannot always ensure success since the feasibility of the original problem can vary between settings. 
% When a setting is infeasible, the robust approach cannot find a solution and so is the nominal one.
%
\section{Conclusions and future work}\label{sec:conclusions}
In this work, we present an uncertainty-aware optimal control formulation that takes the uncertainty in contact events into account using the notion of guards in hybrid systems and enables tractable resolution of the problem. Our proposed formulation features constraint satisfaction and uncertainty optimization within a robust phase, making it applicable to various problems in robotics with uncertain contact events. Several case studies showed that, in addition to generating robust trajectories, uncertainty optimization is important to avoid failure.

In the future, we plan to extend the uncertainty-aware approach to parallelized tree-structured optimal control for applications that emphasize the long-term optimality of multi-phase planning. We also plan to implement a fast parameterized and constrained optimal control solver for real-world experiments. Real-world experiments are also part of our future vision for this work.
%%%%%%%%%%%%%%%%%%%%%%%%%%%%%%%%%%%%%%%%%%%%%%%%%%%%%%%%%%%%%%%%%%%%%%%%%%%%%%%%
\bibliography{master,others} 
\bibliographystyle{ieeetr}

\end{document}